\documentclass{article}

\usepackage{microtype}
\usepackage{graphicx, caption, subcaption}  % sub figures
\usepackage{float}
\usepackage{booktabs} % for professional tables
\usepackage{algorithm}

\usepackage{hyperref}

\usepackage{amsmath}        % extended math environments
\usepackage{amsthm}         % extended math theorems 
\usepackage{amsfonts}       % blackboard math symbols
\usepackage{wasysym}

\usepackage[preprint]{icml2020}

\icmltitlerunning{Fast Fair Regression via Efficient Approximations of Mutual Information}

\newcommand{\A}         {\ensuremath{A}}
\newcommand{\F}         {\ensuremath{f}}
\newcommand{\Y}         {\ensuremath{Y}}
\newcommand{\Sc}        {\ensuremath{S}}
\newcommand{\X}         {\ensuremath{X}}
\newcommand{\x}         {\ensuremath{\mathbf{x}}}
\newcommand{\z}         {\ensuremath{\mathbf{z}}}
\newcommand{\params}    {\ensuremath{\boldsymbol\theta}}
\newcommand{\Ai}        {\ensuremath{a}}
\newcommand{\Yi}        {\ensuremath{y}}
\newcommand{\Sci}       {\ensuremath{s}}
\newcommand{\Ymat}      {\ensuremath{\mathbf{\Yi}}}
\newcommand{\Amat}      {\ensuremath{\mathbf{\Ai}}}
\newcommand{\Scmat}     {\ensuremath{\mathbf{\Sci}}}

\newcommand{\N}         {\ensuremath{n}}
\newcommand{\regw}      {\ensuremath{\lambda}}
\newcommand{\regf}      {\ensuremath{\gamma}}
\newcommand{\regclf}    {\ensuremath{\kappa}}
\newcommand{\clfw}      {\ensuremath{\boldsymbol\Psi}}
\newcommand{\clfwa}     {\ensuremath{\boldsymbol\psi}}

\newcommand{\Loss}      {\ensuremath{\mathcal{L}}}

\newcommand{\As}        {\ensuremath{\mathcal{A}}}

\newcommand{\Ys}        {\ensuremath{\mathcal{Y}}}
\newcommand{\Xs}        {\ensuremath{\mathcal{X}}}

\newcommand{\Scs}       {\ensuremath{\mathcal{S}}}
\newcommand{\paramss}   {\ensuremath{\Theta}}
\newcommand{\real}      {\ensuremath{\mathbb{R}}}
\newcommand{\basisb}    {\ensuremath{\boldsymbol\phi}}
\newcommand{\Basisb}    {\ensuremath{\boldsymbol\Phi}}

\newcommand{\fn}[1]     {\ensuremath{\F\!\left({#1}\right)}}
\newcommand{\pr}[1]     {\ensuremath{\mathrm{P}\!\left({#1}\right)}}
\newcommand{\pc}[2]     {\ensuremath{\mathrm{P}\!\left({#1}|{#2}\right)}}
\newcommand{\loss}[1]   {\ensuremath{\ell_{#1}}}

\newcommand{\mi}        {\ensuremath{\mathrm{I}}}
\newcommand{\nmi}       {\ensuremath{\tilde{\mi}}}
\newcommand{\ent}       {\ensuremath{\mathrm{H}}}
\newcommand{\MI}[2]     {\ensuremath{\mi\!\left[{#1};{#2}\right]}}
\newcommand{\NMI}[2]    {\ensuremath{\nmi\!\left[{#1};{#2}\right]}}
\newcommand{\CMI}[3]    {\ensuremath{\mi\!\left[{#1};{#2}|{#3}\right]}}
\newcommand{\NCMI}[3]   {\ensuremath{\nmi\!\left[{#1};{#2}|{#3}\right]}}
\newcommand{\Ent}[1]    {\ensuremath{\ent\!\left[{#1}\right]}}
\newcommand{\CEnt}[2]   {\ensuremath{\ent\!\left[{#1}|{#2}\right]}}

\newcommand{\clf}[2]    {\ensuremath{\rho\!\left({#1} | {#2}\right)}}
\newcommand{\intd}[1]   {\ensuremath{\mathrm{d}{#1}}}

\newcommand{\indic}[1]  {\ensuremath{\mathbf{1}_{#1}}}
\newcommand{\ident}[1]  {\ensuremath{\mathbf{I}_{#1}}}
\newcommand{\bigo}[1]   {\ensuremath{\mathcal{O}\!\left({#1}\right)}}
\newcommand{\basis}[1]  {\ensuremath{\basisb\!\left({#1}\right)}}

\newcommand{\softplus}[1] {\ensuremath{\mathrm{softplus}({#1})}}

\newcommand{\fit}{\mathrm{fit}}
\newcommand{\fair}{\mathrm{fair}}
\newcommand{\ind}{\mathrm{ind}}
\newcommand{\sep}{\mathrm{sep}}
\newcommand{\suf}{\mathrm{suf}}

\DeclareMathOperator*{\argmin}{arg\!\,min}

\newtheorem*{statement}{Problem Statement}

\begin{document}

\twocolumn[
\icmltitle{Fast Fair Regression via Efficient Approximations of Mutual Information}



\icmlsetsymbol{equal}{*}

\begin{icmlauthorlist}
\icmlauthor{Daniel Steinberg}{gradc}
\icmlauthor{Alistair Reid}{grads}
\icmlauthor{Simon O'Callaghan}{grads}
\icmlauthor{Finnian Lattimore}{grads}
\icmlauthor{Lachlan McCalman}{gradc}
\icmlauthor{Tiberio Caetano}{grads}
\end{icmlauthorlist}

\icmlaffiliation{gradc}{Gradient Institute, Canberra, Australia.}
\icmlaffiliation{grads}{Gradient Institute, Sydney, Australia.}

\icmlcorrespondingauthor{Daniel Steinberg}{dan@gradientinstitute.org}
% \icmlcorrespondingauthor{Alistair Reid}{al@gradientinstitute.org}

\icmlkeywords{fairness, regression, probabilistic classification, mutual information,
entropy, regularization}

\vskip 0.3in
]



\printAffiliationsAndNotice{}  % leave blank if no need to mention equal contribution

\begin{abstract}
%This paper addresses the problem of estimating a real-valued outcome, such as a risk score or insurance premium, under fairness constraints with respect to a protected attribute, such as gender or age group.

Most work in algorithmic fairness to date has focused on discrete outcomes, such as deciding whether to grant someone a loan or not. In these \emph{classification} settings, group fairness criteria such as independence,
separation and sufficiency can be measured directly by comparing rates of outcomes
between subpopulations.
Many important problems however require the prediction of a real-valued outcome, such as a risk score or insurance premium. In such \emph{regression} settings, measuring group fairness criteria is computationally
challenging, as it requires estimating information-theoretic
divergences between conditional probability density functions.
This paper introduces fast approximations of the independence, separation
and sufficiency group fairness criteria for regression models from their
(conditional) mutual information definitions, and uses such approximations as regularisers to enforce fairness within a regularised risk minimisation framework. %
%by showing they can be expressed
%as conditional probabilities of the sensitive attributes that we estimate using
%probabilistic classification.
%
%Based on these fast approximations of mutual information we create regularisers that adjust the
%predictions of regression algorithms towards satisfying these criteria.
%\emph{without} further relaxation.
%
Experiments in real-world datasets indicate that in spite of its superior computational efficiency our algorithm still displays state-of-the-art accuracy$/$fairness tradeoffs.

%, albeit simple and efficient to
%compute, provide an effective means to trade-off accuracy and fairness in
%regression settings.

\end{abstract}

\section{Introduction}%
\label{sec:int}

A machine learning algorithm trained with a standard objective such as maximum
accuracy can produce behaviours that people, including those affected by its
decisions, consider unfair.
Algorithmic group fairness seeks to address this problem by defining protected
attributes such as age, race or gender, and introducing mathematical fairness
measures to enable assessment and potential amelioration of unfair treatment or outcomes.
Many fairness measures have been proposed in the literature, but most can be
posed as specialisations of three general criteria, \emph{independence},
\emph{separation} and \emph{sufficiency}, that can be cleanly and intuitively
defined in terms of statistical independence~\cite{barocas_2019}.

To date, the algorithmic fairness literature has mainly focused on
classification problems, where the decisions of a system
are binary or categorical, such as predicting who should be released on
bail~\cite{propublica_2016}, or who should be audited for a potentially
fraudulent tax return.
In this setting, fairness measures are straightforward to compute based on
the fraction of each group that were correctly or incorrectly selected or
omitted.

However, there are many important problems which involve a real-valued prediction,
such as how much to lend someone on a credit card or home loan,
or how much to charge for an insurance premium. In such continuous settings the aforementioned fairness criteria are generally
intractable to evaluate, and few attempts have been made in this direction~\cite{agarwal_2019,berk_2017,fitzsimons_2019}.

%THE BELOW SEEMS UNSUBSTANTIATED
%Various tractable fairness measures have been proposed, but we believe that
%many are either
%(a) not as intuitive to reason about and thus apply as independence,
%separation and sufficiency, or 
%(b) are simplifications of these that fail to fully capture the important
%properties of the original criteria.

In this paper, we 
\begin{itemize}
\item introduce a technique that produces fast approximations of
    \emph{independence}, \emph{separation} and \emph{sufficiency} group
    fairness criteria in a regression setting based on (conditional) mutual
    information. This technique is \emph{agnostic} to the regression algorithms used.
    % This is achieved by factorising the criteria as conditional probabilities
    % of the (categorical) protected attribute, and apply estimation techniques
    % based on probabilistic classification~\cite{qin_1998} and empirical
    % integration to estimate their values.  
  \item incorporate the resulting approximations as fairness regularisers in a regression setting, thus enabling us to trade-off accuracy and fairness within a regularised risk minimisation framework.
      \item empirically compare our method with a state-of-the-art approach using real data and benchmarking directly
    against these three fairness criteria. To the best of our knowledge we offer the first investigation of the trade-offs between
    accuracy and satisfaction of all of these fairness criteria in a regression setting.
\end{itemize}

%\subsection{Paper Contribution}
%
%We offer the following contributions to the field of algorithmic fairness in
%this paper:
%\begin{enumerate}
%  \item We introduce techniques to tractably approximate the
%    \emph{independence}, \emph{separation} and \emph{sufficiency} group
%    fairness criteria in a regression setting based on (conditional) mutual
%    information.
%    These techniques are \emph{agnostic} to the regression algorithms used.
%    % This is achieved by factorising the criteria as conditional probabilities
%    % of the (categorical) protected attribute, and apply estimation techniques
%    % based on probabilistic classification~\cite{qin_1998} and empirical
%    % integration to estimate their values.  
%  \item Based on these approximations, we present two simple regularisation
%    algorithms that adjust the predictions of a regression algorithm towards
%    satisfying these criteria \emph{without} requiring these criteria to be
%    simplified.
%  \item We compare these regularisation techniques to other in-processing
%    techniques in the literature, using real data and benchmarking directly
%    against these three fairness criteria.
%    To the best of our knowledge this is the first time in a regression setting where the trade-off between
%    accuracy and satisfaction of all of these fairness criteria has been empirically analysed.
%\end{enumerate}
%

\subsection{Related Work}

Predictive fairness measures date back more than fifty years in the context
of testing and hiring, as summarised by~\citet{hutchinson_2019}.
Of particular relevance to our work are the measures of regression fairness
based on (partial) correlations between the predicted score, the target
variable and the demographic group of applicants, as derived
in~\cite{cleary_1968, darlington_1971}.
These measures can be interpreted as relaxations of the independence,
separation and sufficiency criteria under 
the assumption that the outcome, prediction and group are jointly Gaussian distributed.
Propensity score stratification techniques have also been used to 
decrease the separation of the protected class for linear
regressors in~\cite{calders_2013}.
\citet{berk_2017} propose fairness objectives that address separation using
convex regularisers.
One of these regularisers (`group') permits cancelling of model errors within a
group, addressing an imbalance in expectation, while the other (`individual')
does not.
\citet{fitzsimons_2019} have developed an inprocessing technique for kernel
machines and decision tree regressors with a focus on satisfying (a relaxation
of) the independence criterion. \citet{agarwal_2019} propose methods for fair regression of real-valued outputs for both the independence criterion and a bounded group loss criterion. For bounded group loss the approach is based on weighted loss minimisation, whereas for the independence criterion the algorithm proposed involves a reduction to classification based on a suitable discretisation of the real-valued output space. 
\citet{williamson_2019} proposed that risk measures from mathematical
finance generalise to fairness measures by imposing that the distribution of
losses across subgroups are commensurate.
Of particular interest, the conditional value at risk measure leads to a convex
inprocessing technique for both regression and classification problems.

Mutual information (MI) has also previously been considered in the context of fair
supervised learning.
\citet{kamishima_2012} use normalised MI to asses fairness in
their \emph{normalised prejudice index} (NPI).
Their focus is on binary classification with binary sensitive attributes, and
the NPI is based on the independence fairness criterion.
In such a setting mutual information is readily computable empirically from
confusion matrices.
This work is generalised in~\cite{fukuchi_2015} for use in regression models by
using a \emph{neutrality} measure, which the authors show is equivalent to the
independence criterion.
They then use this neutrality measure to create inprocessing techniques for
linear and logistic regression algorithms.
Similarly,~\citet{ghassami_2018} take an information theoretic approach to
creating an optimisation algorithm that returns a predictor score that is fair
(up to some $\epsilon$) with respect to the separation criterion. Though
they have left as future work to test an implementation of this algorithm
on real data.
In constrast, the focus in our paper is to present fast and scalable methods of
approximating conditional mutual information specifically tailored to encoding
the independence, separation and sufficiency criteria within a learning
algorithm.

\section{Problem Formulation}

We consider a predictor with target set, $\Ys$, feature
set, $\Xs$, and a sensitive attribute set, $\As$ and corresponding random variables $\Y$, $\A$ and $\X$. Our data is assumed drawn from some
distribution over $\Ys \times \Xs$
($\As$ is a subset of $\Xs$).
Here $\F$ is a prediction function $\F : \Xs \to \Scs$ that produces scores
$\Sc = \fn{\X, \params}$, with parameters $\params \in \paramss$.
For the purposes of this paper we will treat $\Sc$ as a random variable drawn
from a distribution over~$\Scs$, and also that $\Scs = \Ys$.

We are specifically interested in approximating \emph{independence}, 
\emph{separation} and \emph{sufficiency}, a selection that subsumes
many other measures as relaxations or special cases~\cite{barocas_2019}. These
can be defined in terms of independence or conditional independence statements,
\begin{align}
  &\text{Independence:} \nonumber \\
  &\qquad \Sc \bot \A \Leftrightarrow \pr{\Sc, \A} = \pr{\Sc}\pr{\A},
    \label{eq:ind} \\
  &\text{Separation:} \nonumber \\
  &\qquad \Sc \bot \A~|~\Y \Leftrightarrow \pc{\Sc, \A}{\Y} = \pc{\Sc}{\Y}\pc{\A}{\Y}, 
    \label{eq:sep} \\
  &\text{Sufficiency:} \nonumber \\
  &\qquad \Y \bot \A~|~\Sc \Leftrightarrow \pc{\Y, \A}{\Sc} = \pc{\Y}{\Sc}\pc{\A}{\Sc}. 
    \label{eq:suf}
\end{align}
%Here we have simplified notation by overloading $\mathrm{P}$.
Strictly speaking, all of these distributions are different functions
that take instances of random variables as input, e.g.
$\mathrm{P}_{\A | \Sc}(\A=\Ai | \Sc=\Sci)$.
In the interest of being concise, we will assume the reader can infer
the specific probability distribution given either random variable inputs
in the generic case, $\pc{\A}{\Sc}$, or specific instances of them, $\pc{\Ai}{\Sci}$,
where appropriate.

Rather than constraining the predictor such that these conditions be exactly
satisfied, we would usually define a continuous measure of the degree to which
they are satisfied. This admits a balancing of priorities between these and other
measures such as accuracy.

In the \emph{classification} setting where $\Y$, $\Sc$ and $\A$ are all
categorical, these conditional probabilities can be empirically estimated from
confusion matrices between $\Y$ and $\Sc$ for each protected subgroup
~\cite{barocas_2019}.
In a regression setting, however, they become continuous density functions,
and we are required to either assume a parametric form for them, or numerically
estimate the densities using methods such as kernel density
estimation~\cite{hastie2001}. Others have simplified these criteria using conditional expectation instead
of conditional probability~\cite{calders_2013, zafar_2015, fitzsimons_2019}.
This approach however can fail to capture important effects that can still lead to harm, such as groups with
different score variances.

Our primary aim in this paper is to create techniques that tractably approximate the three above
criteria of group fairness in a way that is agnostic to the
predictor used to generate the score, $\Sc$, and that can be incorporated 
into a regression learning objective. More formally,

\begin{statement}

  Derive methods that can approximate the fairness criteria
  \emph{independence}, \emph{separation} and \emph{sufficiency}, for $\Ys
  \subseteq \real$ and $\Scs \subseteq \real$, in the case of \emph{binary}
  or \emph{categorical} sensitive attributes, $\As = \{0, \ldots,
  K\}$, $K\ge 1$, and incorporate the resulting approximations as regularisers into a regression
  learning objective.

\end{statement}

\section{Measuring Group Fairness}%
\label{sec:mi}

A natural approach to measuring the fairness criteria
in~\eqref{eq:ind}-\eqref{eq:suf} is to calculate the (conditional) mutual
information (MI)~\cite{gelfand_1957, cover2006} between the relevant variables.
For instance, to asses the \emph{independence} criterion from~\eqref{eq:ind},
we can calculate the MI between $\Sc$ and~$\A$~\cite{fukuchi_2015},
\begin{align}
  \MI{\Sc}{\A} = \int_\Scs \sum_{\Ai \in \As} \pr{\Sci, \Ai} 
  \log \frac{\pr{\Sci, \Ai}}{\pr{\Sci}\pr{\Ai}} \intd{\Sci}.
  \label{eq:mi_ind}
\end{align}
Here we can see that if we achieve perfect
fairness as defined by independence, we have $\pr{\Sc, \A} = \pr{\Sc}\pr{\A}$, which implies $\MI{\Sc}{\A} = 0$; otherwise MI
will be positive\footnote{This can be shown by applying Jensen's Inequality 
to~\eqref{eq:mi_ind}.}.
We can also \emph{normalise} MI by one of its many known upper bounds so that
it takes values in $[0, 1]$. 
One useful upper bound in this context is the entropy of the sensitive
attribute, $\Ent{\A}$, that gives the normalised measure,
\begin{align}
  \NMI{\Sc}{\A} = \frac{\MI{\Sc}{\A}}{\Ent{\A}}, ~~ \text{where} ~
  \Ent{\A} = - \sum_{\Ai \in \As} \pr{\Ai} \log \pr{\Ai}.
  \label{eq:nmi_ind}
\end{align}
The reason for using $\Ent{\A}$ as a normaliser is because MI and entropy are
related through $\MI{\Sc}{\A} = \Ent{\A} - \CEnt{\A}{\Sc}$, where
$\CEnt{\A}{\Sc}$ is conditional entropy, and is a measure of how much of the
information of $P(A)$ is encoded by $P(S)$.
When $\NMI{\Sc}{\A} = 1$ we have that $\CEnt{\A}{\Sc} = 0$, so the distribution of
$\Sc$ completely encodes all information about $\A$.
This would be \emph{maximally} unfair according to the independence criterion of fairness,
since we can completely recover all information about the sensitive attribute
from the model's predictions.

Since we do not have access to these probability densities we have to resort to
approximating \NMI{\Sc}{\A} from samples.
Many general approximation methods for mutual information exist in the
literature, for example the popular ``KSG'' method~\cite{kraskov_2004} and
improvements~\cite{gao_2018}, and recent work including~\cite{belghazi_2018}.
When considering categorical protected attributes in a regression context,
methods that can handle mixed categorical/continuous distributions are
necessary, for example~\cite{gao_2017}, which we do compare against. 
However, these methods have runtime performance and computational complexity
that makes them unsuitable for use in a regularisation term during training. 
This motivates our proposal of a \emph{fast} approximation method that handles the mixed categorical/continuous case resulting from a discrete protected attribute and a continuous, real-valued outcome variable.

We start by noting that $\Ent{\A}$ is easy to approximate empirically,
\begin{align}
  \Ent{\A} \approx - \sum_{\Ai \in \As} \frac{\N_\Ai}{\N} \log \frac{\N_\Ai}{\N},
  \label{eq:ah_ind}
\end{align}
where $\N_\Ai$ is the number of instances that have sensitive attribute $\A = \Ai$,
and $\N$ is the total number of instances in the dataset.

In order to approximate $\MI{\Sc}{\A}$, note it can be rewritten as% $\MI{\Sc}{\A}$ as
\begin{align}
  \MI{\Sc}{\A} = \int_\Scs \sum_{\Ai \in \As} \pr{\Sci, \Ai}
    \log \frac{\pc{\Ai}{\Sci}}{\pr{\Ai}} \intd{\Sci}
\end{align}
We can again empirically estimate $\pr{\Ai} \approx \N_\Ai / \N$, and since
$\Ai$ is categorical we can use class probability estimation to estimate
$\pc{\Ai}{\Sci}$ directly,
\begin{align}
  \pc{\A = a}{\Sc=\Sci} &\approx \clf{a}{\Sci}.
\end{align}
Here $\clf{\Ai}{\Sci}$ is a prediction of the probability that class $\A=a$.
%This is a well known trick that is commonly used in the context of density
%ratio estimation~\cite{qin_1998, bickel_2009, sugiyama_2010}. 
If we then estimate the integral--sum over $\Scs \times \As$ empirically
we obtain the simple approximation to MI
\begin{align}
   \MI{\Sc}{\A} \approx \frac{1}{\N} \sum^\N_{i=1}
    \log \frac{\clf{\Ai_i}{\Sci_i}}{\N_{a_i} / \N},
  \label{eq:ami_ind}
\end{align}
where $\Ai_i$ is the sensitive class of instance $i$. 
Finally we can combine~\eqref{eq:ah_ind} and~\eqref{eq:ami_ind} to approximate
$\NMI{\Sc}{\A}$ as defined in~\eqref{eq:nmi_ind}.

We can proceed along similar lines for the separation and sufficiency criteria,
though these require conditional mutual information with conditional entropy
normalisers,
\begin{align}
  \text{Separation:} \qquad
  \NCMI{\Sc}{\A}{\Y} &= \frac{\CMI{\Sc}{\A}{\Y}}{\CEnt{\A}{\Y}}, \nonumber \\
  \text{Sufficiency:} \qquad
  \NCMI{\Y}{\A}{\Sc} &= \frac{\CMI{\Y}{\A}{\Sc}}{\CEnt{\A}{\Sc}}.
  \label{eq:nmi_ss}
\end{align}
Here we can approximate the conditional entropy normalisers again using probabilistic
classifiers and estimating the integrals empirically,
\begin{align}
  \CEnt{\A}{\Y} =& - \int_\Ys \sum_{\Ai \in \As} \pr{\Yi, \Ai} 
    \log \pc{\Ai}{\Yi} \intd{\Yi}, \nonumber \\
  \approx& - \frac{1}{\N} \sum_{i=1}^{\N} \log \clf{\Ai_i}{\Yi_i}, \nonumber \\
  \CEnt{\A}{\Sc} =& - \int_\Scs \sum_{\Ai \in \As} \pr{\Sci, \Ai}
    \log \pc{\Ai}{\Sci} \intd{\Sci}, \nonumber \\
  \approx& - \frac{1}{\N} \sum_{i=1}^{\N} \log \clf{\Ai_i}{\Sci_i}.
\end{align}
These conditional entropy normalisers play a similar role as entropy in the
normalised independence criterion measure.
For separation and sufficiency, if respectively $\NCMI{\Sc}{\A}{\Y} = 1$ or
$\NCMI{\Y}{\A}{\Sc} = 1$, this means that $\CEnt{\A}{\Sc, \Y} = 0$
in both cases.
Intuitively, we can interpret this to mean that \emph{jointly} $\Y$ and $\Sc$
totally determine $\A$, which is by definition maximally unfair according to
both of these fairness criteria when $\CEnt{\A}{\Y} > 0$ and $\CEnt{\A}{\Sc} >
0$ respectively.
Using the class probability estimates
$\pc{\Ai}{\Yi} \approx \clf{\Ai}{\Yi}$ and $\pc{\Ai}{\Yi, \Sci} \approx
\clf{\Ai}{\Yi, \Sci}$, we can estimate the conditional mutual information entailed by the
separation criterion,
\begin{align}
  \CMI{\Sc}{\A}{\Y} &= \iint\limits_{\Ys~\Scs} \sum_{\Ai \in \As}
    \pr{\Yi, \Sci, \Ai} \log \frac{\pc{\Ai}{\Yi, \Sci}}{\pc{\Ai}{\Yi}}
    \intd{\Yi} \intd{\Sci} \nonumber \\
  &\approx \frac{1}{\N} \sum_{i=1}^{\N}
    \log \frac{\clf{\Ai_i}{\Yi_i, \Sci_i}}{\clf{\Ai_i}{\Yi_i}},
\end{align}
as well as by the sufficiency criterion,
\begin{align}
  \CMI{\Y}{\A}{\Sc} &= \iint\limits_{\Ys~\Scs} \sum_{\Ai \in \As}
    \pr{\Yi, \Sci, \Ai} \log \frac{\pc{\Ai}{\Yi, \Sci}}{\pc{\Ai}{\Sci}}
    \intd{\Yi} \intd{\Sci} \nonumber \\
  &\approx \frac{1}{\N} \sum_{i=1}^{\N}
    \log \frac{\clf{\Ai_i}{\Yi_i, \Sci_i}}{\clf{\Ai_i}{\Sci_i}}.
\end{align}
% Note that the above criteria differently combine the outputs of three distinct
% probabilistic classifiers separately predicting \clf{\Ai}{\Sci_i},
% \clf{\Ai}{\Yi_i, \Sci_i} and \clf{\Ai}{\Yi_i} and use the occurrence
% frequency to estimate the scalar quantity \pr{\Ai}.

%We will refer to these approximated versions of the normalised (conditional) mutual information
%criteria in~\eqref{eq:nmi_ind} and~\eqref{eq:nmi_ss} as $\ami{ind}$,
%$\ami{sep}$, and $\ami{suf}$ respectively. These are the key quantities we will operate with.

%As stated in the literature review, there are a plethora of methods available
%for estimating mutual information.
%However, as noted by \citet{gao_2017} these methods do not usually work well
%with mixed types of variables (continuous-categorical).
%The class-conditional probability estimation outlined above is specific
%to this scenario, and would not work for general mutual information
%estimation.

Given the above derivations, our task now is to come up with estimators for the quantities $\clf{\Ai}{\Sci}$ $\clf{\Ai}{\Yi}$ and $\clf{\Ai}{\Yi, \Sci}$ and incorporate the resulting estimates of mutual information as regularisers in a regression learning objective. That's the focus of the next section.

\section{Learning with Fairness Regularisers}%
\label{sec:reg}

%In this section we present two estimators for the class-conditional probabilities outlined in the previous section, which result in two mutual information estimators which in turn give rise to two types of fairness regularisers. 

Before presenting our estimator for $\clf{\Ai}{\Sci}$, $\clf{\Ai}{\Yi}$ and $\clf{\Ai}{\Yi, \Sci}$, we describe how these quantities can be used to construct fairness entropic regularisers in a learning objective.

\subsection{Fairness Regularisers}

For simplicity we will assume we can express the regressor of interest as $\Sci
= \fn{\x, \params}$. Then learning the optimal score can be cast as
(empirically) minimising a fitting loss, $\loss{\fit}$, and a regularisation
loss, $\loss{\params}$, with respect to model parameters, $\params \in
\paramss$, i.e.,
\begin{align}
  \params^* = \argmin_{\params \in \paramss} 
   \loss{\fit}(\Ymat, \Scmat)
   + \regw\loss{\params}(\params),
\end{align}
where $\regw$ is a weighting coefficient applied to the parameter regulariser
(chosen by cross validation), and bold denotes a vector of instances, $\Ymat =
[\Yi_i, \dots, \Yi_\N]^\top$, $\Scmat= [\Sci_i, \dots, \Sci_\N]^\top$.
What we aim to do in this section is introduce
an additional regularizer, $\loss{\fair}$, that adjusts the predictions
towards satisfying a particular fairness criterion,
\begin{align}
  \params^*= \argmin_{\params \in \paramss}
  \loss{\fit}(\Ymat, \Scmat)
  + \regw\loss{\params}(\params)
  + \regf \loss{\fair}(\Ymat, \Scmat, \Amat),
  \label{eq:loss_fair}
\end{align}
where $\Amat = [\Ai_i, \dots, \Ai_\N]^\top$ and $\regf$ is the
fairness regulariser weight.
The most natural starting point is to incorporate the normalised (conditional)
mutual information measures, $\nmi$, into $\loss{\fair}$. The intuition is simple: enforcing fairness means penalising deviations from the ideal fairness criteria in eqs. \eqref{eq:ind}, \eqref{eq:sep} and \eqref{eq:suf}, and these are quantified by $\nmi$.
However we only require functions that share
the same minimum as $\nmi$ with respect to $\params$.
Using the identities $\MI{U}{V} = \Ent{V} - \CEnt{V}{U}$ and $\CMI{U}{V}{W} = \CEnt{V}{W} - \CEnt{V}{U, W}$ we obtain
\begin{align}
  \min_{\params} \MI{\Sc}{\A} &= \min_{\params} - \CEnt{\A}{\Sc} \label{eq:ind_lb}\\
  \min_{\params} \CMI{\Sc}{\A}{\Y} &= \min_{\params} - \CEnt{\A}{\Y, \Sc} \label{eq:sep_lb} \\
  \min_{\params} \CMI{\Y}{\A}{\Sc} &= \min_{\params} \CEnt{\A}{\Sc} - \CEnt{\A}{\Y, \Sc} \label{eq:suf_lb},
\end{align}
where we have dropped all terms independent of $\params$.
%Interestingly, it appears that the sufficiency regulariser in~\eqref{eq:suf_lb} can be expressed as a linear combination of the other two regularisers (\todo could this suggest that there are only really two fairness criteria?).
Now we can again employ the same tricks we have used previously to approximate
the entropies in~\eqref{eq:ind_lb}-\eqref{eq:suf_lb} to arrive at expressions for fairness regularisers,
\begin{align}
  \loss{\ind}(\Scmat, \Amat)
  =& \frac{1}{\N} \sum_{i=1}^\N \log \clf{\Ai_i}{\z_{\ind, i}, \clfw}
    \label{eq:reg_ind} \\
  \loss{\sep}(\Ymat, \Scmat, \Amat)
  =& \frac{1}{\N} \sum_{i=1}^\N \log \clf{\Ai_i}{\z_{\sep, i}, \clfw}
    \label{eq:reg_sep} \\
  \loss{\suf}(\Ymat, \Scmat, \Amat)
%    =& \log \clf{\Ai}{\z_\sep}
%     - \log \clf{\Ai}{\z_\ind}, \label{eq:reg_suf}
    =&~\loss{\sep}(\Ymat, \Scmat, \Amat)
     - \loss{\ind}(\Scmat, \Amat), \label{eq:reg_suf}
\end{align}
where $\clfw$ are classifier parameters and we define
\begin{align}
  \z_\ind = [1, \Sci]^\top \quad \text{and} \quad
  \z_\sep = [1, \Sci, \Yi]^\top \label{eq:z},
\end{align}
with 1 included as an optional intercept/offset term.
We then incorporate~\eqref{eq:reg_ind},~\eqref{eq:reg_ind} or~\eqref{eq:reg_suf} into the sum 
in~\eqref{eq:loss_fair} for $\loss{\fair}$ (depending on which fairness criterion we want to enforce).

The final step is to find computationally attractive means of
estimating the probabilistic classifiers, $\clf{\Ai}{\cdot}$, in the
optimisation procedure for~\eqref{eq:loss_fair}. Naively
we could use a logistic regression (or softmax) classifier for this task,
but this would require us to run an iterative optimisation 
to learn the weights of such a classifier for every iteration
\emph{within} the optimisation procedure of~\eqref{eq:loss_fair}.
To circumvent this issue, we present an alternative, more
computationally efficient method\footnote{In the supplement we show 
results with a logistic regression classifier}.

\subsection{Least Squares Probabilistic Classification}%
\label{sub:LSPC}

The approach we introduce for estimating the probabilistic
classifiers in~\eqref{eq:reg_ind} and~\eqref{eq:reg_sep} consists of applying the idea of least squares
probabilistic classification (LSPC)
~\cite{sugiyama_efficient_2010, sugiyama_superfast_2010}. 
%For a detailed derivation and analysis of the method we refer the reader
%to the two aforementioned papers. 
Using generic inputs $\z \in \real^p$, LSPC approximates posterior probabilities of the sensitive class, $\A$, as
\begin{align}
  [\clf{1}{\z}, \ldots, \clf{K}{\z}]^\top = \clfw\basis{\z}.
\end{align}
Here $\basisb: \real^p \to \real^D$ is a (nonlinear) basis function, and
$\clfw \in \real^{K \times D}$ are the classifier weights that are learned
using a regularised least squares objective,
\begin{align}
  \clfw^* = \argmin_{\clfw \in \real^{D \times K}}
  \frac{1}{\N} \sum^\N_{i=1} \| \indic{\Ai_i} - \clfw\basis{\z_i} \|^2_2
  + \regclf \| \clfw \|^2_\textrm{F},
\end{align}
where $\indic{a} : a \to \{0,
1\}^K$ is a categorical indicator function, $\regclf$ is a regulariser coefficient and $\|\cdot\|_\textrm{F}$ is
the Frobenius norm. 
This optimisation problem has the analytical solution
\begin{align}
  \clfw^* = \left(\Basisb^\top\Basisb + \regclf\ident{D} \right)^{-1}
    \Basisb^\top\indic{\Amat}, \label{eq:clfwopt}
\end{align}
where $\Basisb = [\basis{\z_1}; \ldots; \basis{\z_\N}]$, 
$\indic{\Amat} = [\indic{\Ai_1}; \ldots; \indic{\Ai_\N}]$
(semicolons denoting vertical
concatenation) and $\ident{D}$ is the $D \times D$ identity matrix.
%The computational cost of solving for $\clfw^*$ is dominated by the matrix
%multiplications involving $\Zmat$ with $\bigo{\N D^2}$ and $\bigo{\N D K}$, if
%$\N > D$.
This approximation of posterior class probabilities can be shown to converge
to the true posterior class probabilities,
though a small sample correction procedure is recommended~\cite{kanamori_2010, sugiyama_superfast_2010}.
We have modified this recommended procedure to produce smooth gradients
and to work with input into logarithmic functions. As such our procedure consists of adjusting the class conditional probabilities as follows
\begin{align}
  \clf{\Ai}{\z} = \frac{\softplus{\clfwa_\Ai^\top \basis{\z}}}
  {\sum^K_{k=1} \softplus{\clfwa_k^\top \basis{\z}}}, \label{eq:lspc}
\end{align}
where $\clfwa_k$ is the $k$th row of $\clfw$, and we have replaced $\max\{0,
t\}$ from the original procedure with $\softplus{t} = \log(1 + e^{\beta t}) /
\beta$ that approaches $\max$ as $\beta \to \infty$.
In order to implement the probabilistic estimates
in~\eqref{eq:reg_ind}-\eqref{eq:reg_suf} we just need to substitute the
appropriate $\z$ from~\eqref{eq:z} into the above.

We have to be careful of our choice of basis function, $\basisb$, since
this will directly affect the complexity of learning the model parameters,
$\params$.
Using the identity basis function is appealing in this regard as it leads to
simple and efficient gradient computations. 
This must be traded off against the bias introduced in approximating the
conditional entropy fairness regulariser, as it cannot model non-linear
relationships between the score, target and sensitive attribute.
Another choice of basis function we found to be effective and efficient
was a feature-cross basis that can model quadratic relationships.
%

%Unfortunately this regulariser is not a convex function with respect to
%$\params$, even if the prediction model is linear, $\fn{\x, \params} =
%\params^\top\x$. Therefore we cannot make any global convergence guarantees.
%In practice though we find the algorithm quickly converges to a local optimum.
%\todo Tiberio can perhaps tidy this statement up!

\subsection{Learning with Fairness Regularisers}%
\label{sub:fairreg}

The complete algorithm is sketched out in Algorithm~\ref{alg:lspc}. 
It takes as input the data, regularisation parameters and the
fairness criterion used for regularisation (independence, separation or
sufficiency). 
Each gradient descent step of the model parameters, $\params$, requires
computing the estimate for the fairness regulariser, and we use L-BFGS~\cite{byrd_1995} for the gradient descent algorithm.
% Each iteration alternates between computing the estimate for the fairness
% regulariser (through LSPC and the softplus adjustment), then updating the
% model's parameters based on that estimate.

% We use the L-BFGS algorithm~\cite{byrd_1995} to update the model parameters
% within each iteration.
% Note that the optimisation problem is not convex since the fairness
% regularisers are themselves not convex. 

\begin{algorithm}[tb]
   \caption{LSPC-based optimisation routine for $\params^*$}%
   \label{alg:lspc}
\begin{algorithmic}
  \STATE{\bfseries Input:} data $\{\x_i, \Yi_i, \Ai_i\}_{i=1}^\N$,
    coefficients $\regw$, $\regf$, $\regclf$, fairness criterion $\fair\in\{\ind,\sep,\suf\}$
  \vspace{0.5em}
  \STATE $\params_0 \gets \mathrm{init\_params}()$
  \vspace{0.2em}
  \FUNCTION{$\mathrm{Loss}(\params)$}
    \STATE $\{\Sci_i\}^\N_{i=1} \gets \{\fn{\x_i, \params}\}^\N_{i=1}$
    \STATE $\{\z_i\}_{i=1}^\N \gets$ Equation~\eqref{eq:z}
    \STATE $\clfw^* \gets$ Equation~\eqref{eq:clfwopt}
    \STATE $\{\clf{\Ai_i}{\z_i, \clfw^*}\}^\N_{i=1} \gets$ Equation~\eqref{eq:lspc} 
    \STATE $\loss{\fair}\gets$ Equation~\eqref{eq:reg_ind},~\eqref{eq:reg_sep} or 
      \eqref{eq:reg_suf}
    \STATE $\Loss \gets 
      \loss{\fit}(\Ymat, \Scmat)
      + \regw\loss{\params}(\params)
      + \regf\loss{\fair}(\Ymat, \Scmat, \Amat) $
    \STATE{\bfseries return:} $\Loss$
  \ENDFUNCTION
  \vspace{0.2em}
  \STATE $\mathrm{Gradient} \gets \mathrm{AutoDiff}(\mathrm{Loss})$
  \STATE $\params^* \gets \mathrm{L\text{-}BFGS}(\params_0, \mathrm{Loss},  
    \mathrm{Gradient})$
  \vspace{0.2em}
  \STATE{\bfseries Return:} $\params^*$
\end{algorithmic}
\end{algorithm}

\begin{figure*}[ht]%
\centering
\begin{subfigure}[t]{\linewidth}
  \centering
  \includegraphics[width=\linewidth]{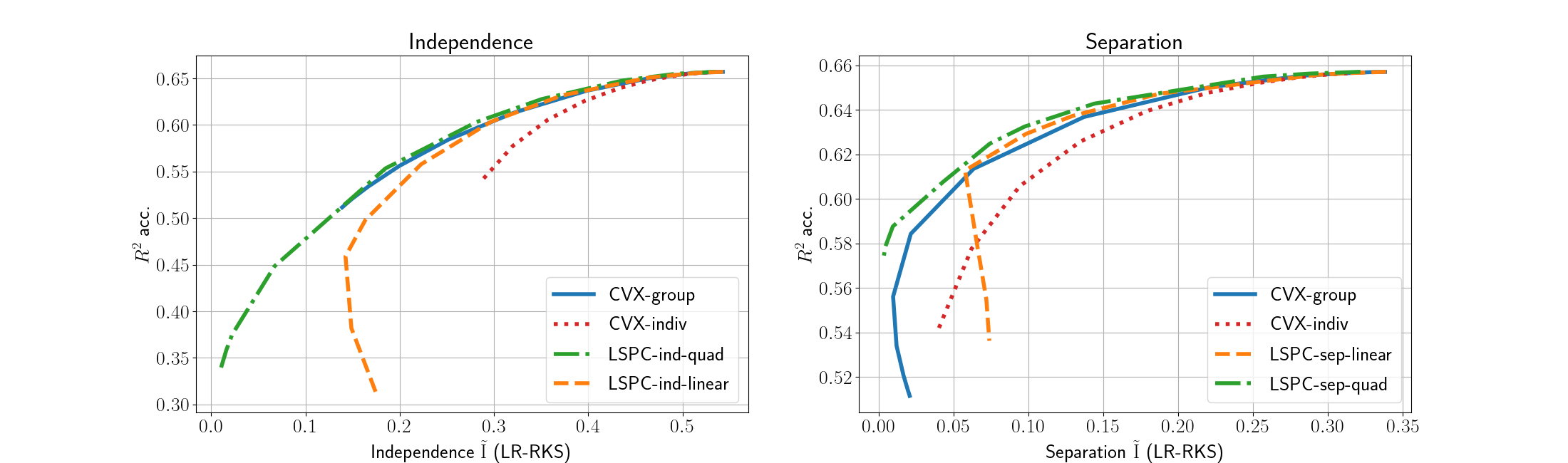}
  \caption{Communities and Crime dataset.\label{fig:reg_comm}
  }
\end{subfigure} \\
\begin{subfigure}[t]{\linewidth}
  \centering
  \includegraphics[width=\linewidth]{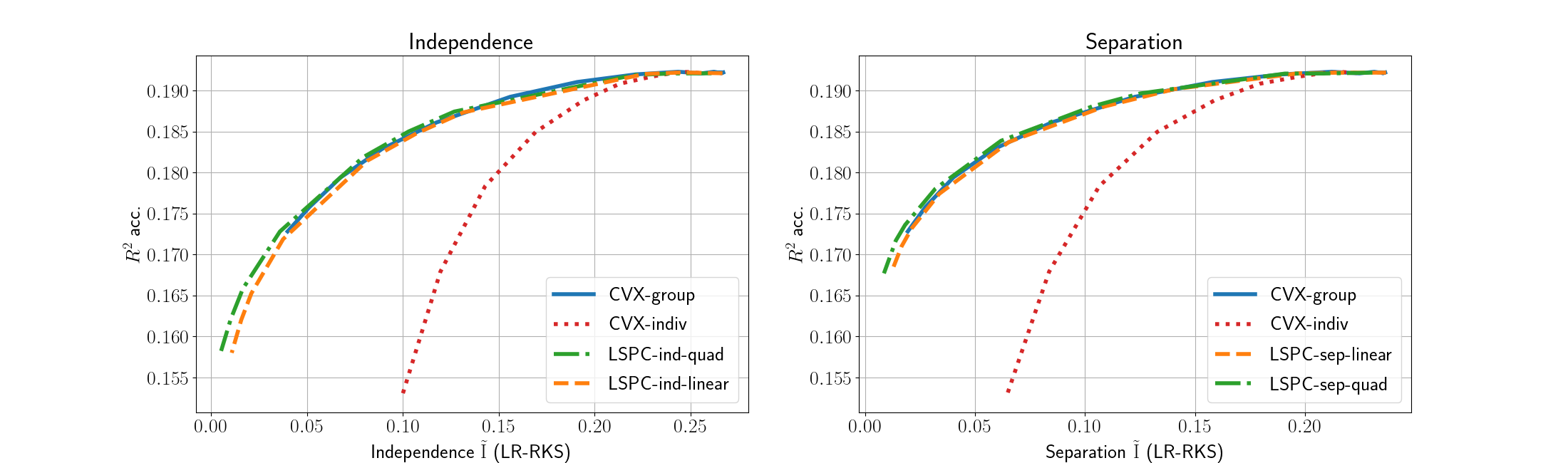}
  \caption{LSAC dataset.\label{fig:reg_lsac}}
\end{subfigure}
\caption{
  Efficiency comparison between our proposed methods, and the convex
  regularises proposed in~\citet{berk_2017}. CVX-group is the `group' convex
  regulariser, and CVX-indiv the `individual'. LSPC are our proposed methods,
  with `ind' and `sep' denoting the independence and separation specific
  regularisers, and `linear' and `quad' denoting the linear or feature-cross
  basis, $\basis{\z}$, respectively. We give an explanation of these results in
  the text.
}
\end{figure*}

%=======
%\subsection{Kernel Probability Estimation}
%\label{sub:RBF}
%\todo Al
%  

\section{Experiments}%
\label{sec:exp}
We start this section by comparing two variants of the proposed approach against two variants of a state-of-the-art method that has the same structure as ours: regularised risk minimisation with a fairness regulariser. We make sure that the \emph{only} difference in the optimisation problems are the fairness regularisers - i.e., the fitting loss is the same and the regularisation on the parameters is also the same. This is to ensure a fair comparison between the two approaches. We also conduct a runtime experiment which showcases how the two approaches scale with a growing sample size. Next we present an empirical study showing how our fast approximation to mutual information compares in terms of accuracy against more computationally expensive approximations.

%This section seeks to answer two key questions. Specifically;
%\begin{itemize}
%    \item how efficiently does the proposed method trade-off accuracy with fairness compared to other methods (Sec.~\ref{sub:exp_reg})? 
%    \item how well does the proposed regularisers approximate the mutual information measure (Sec.~\ref{sub:exp_mi})?
%\todo Update the first point.
%\end{itemize}
%=======
%This section seeks to answer two key questions. Specifically;
%\begin{itemize}
%    \item how efficiently does the proposed method trade-off accuracy with fairness compared to other methods (\S\ref{sub:exp_reg})? 
%    \item how well does the proposed regularisers approximate the mutual information measure (\S\ref{sub:exp_mi})?
%\todo Update the first point.
%\end{itemize}
%>>>>>>> 8268069d5706cd74a614b67cb5afe5c0b7da3a5d

\subsection{Entropy Regularisers}%
\label{sub:exp_reg}

In this experiment we will evaluate how well our fairness regularisers
from \S\ref{sec:reg} perform on real data.
The base model we use for these experiments is a linear model using squared
error loss, $\loss{\fit} = \frac{1}{\N} \sum_{i=1}^\N \| \Yi_i -
\params^\top\x_i\|^2_2$, with $\ell_2$-regularisation, $\loss{\params} =
\|\params\|^2_2$.

We use two variants for $\basis{\z}$; the first being the identity basis (resulting in a linear classifier), and the second a feature-cross basis (resulting in a quadratic nonlinearity).

The convex regularisers, `group' and `individual', proposed by
\citet{berk_2017} are also compatible with this base model configuration.
These regularisers are not explicitly trying to satisfy any of the fairness
criterion we are interested it, but they are similar to the separation
criterion.
The inprocessing method presented by~\citet{komiyama_2018} is also similar to
these regularisers, however they use a different base model ($\x$ are
decorrelated from $\Ai$), and so we have left their approach out of our
comparison.
We have compared the regularisers based on how efficient they are at satisfying
the fairness criteria and accuracy, as measured by $R^2$, for different
regulariser coefficient $\regf$ strengths \emph{ceteris paribus}.
All of these experiments used 5-fold cross validation, and all measures
were computed on held out data (including those for $\nmi$).

\begin{figure}[ht]%
  \includegraphics[width=\linewidth]{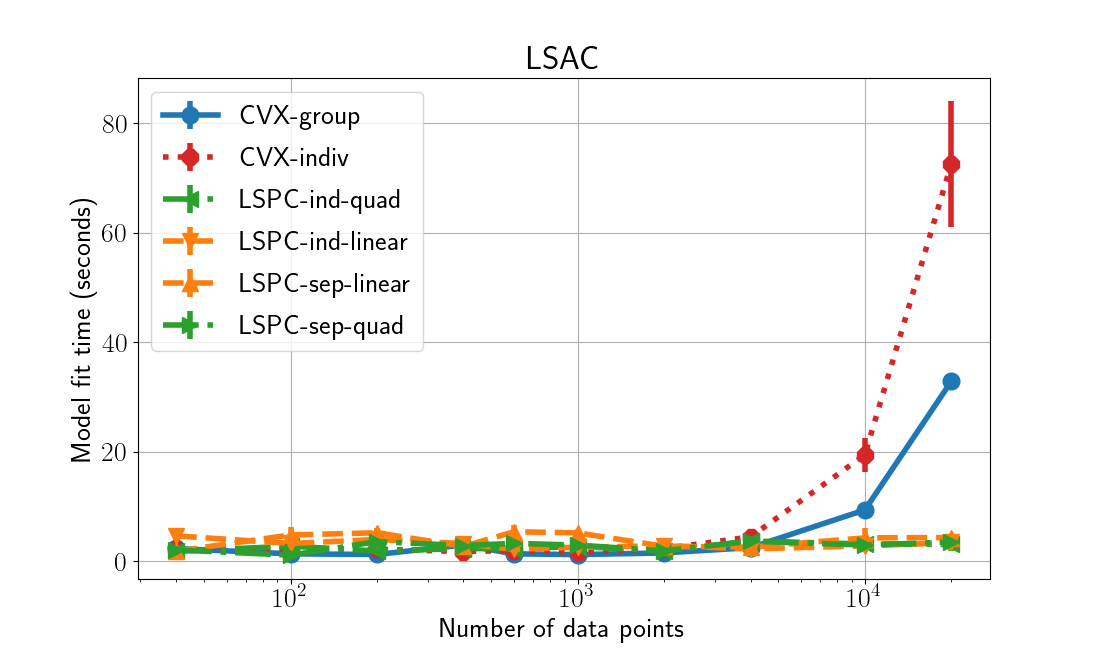}
  \caption{
    Timing comparison between our proposed regularisers and those proposed
    in~\citet{berk_2017}. The x-axis denotes the subset size of the LSAC
    dataset used to time the model fitting, and the bars depict the range 
    of times for a sweep of regulariser strengths,~$\regf$.
  }%
  \label{fig:reg_time}
\end{figure}

The first dataset we use for this comparison is the Communities and crime
dataset (CC)~\cite{redmond_2002, Dua_2019}.
The task is to predict the number of reported violent crimes for 1994
communities across the United States.
Each community instance contains 128 demographic features from the census such
as population density, average income and percentage of population that is
unemployed.
We have identified race as a sensitive attribute, and communities where more
than 50\% of the population identified as Black were labelled as protected as
in~\cite{berk_2017}.

The next dataset is the Law school admission council (LSAC) national
longitudinal bar passage study~\cite{wightman_1998}.
The task is to predict student GPA from a number of demographic and academic
history based features.
Again race is the sensitive class.
We have chosen a random subset of 5000 instances (from more than 20,000) for
the comparison so we could perform an exhaustive parameter sweep of all of the
models.
We found this number did not drastically reduce base-model predictive
performance.

The efficiency results are summarised in Figures~\ref{fig:reg_comm}
and~\ref{fig:reg_lsac}. We have \emph{not} included the sufficiency
results as predictions from the base linear models already satisfied
this criterion, making the comparison uninformative.

In the CC results (Figure~\ref{fig:reg_comm}) we can see that the 
LSPC regularisers with the quadratic (feature-cross) basis were the 
most efficient at satisfying the fairness criteria. 
The `group' convex regulariser was also similarly efficient for some settings
of the regulariser coefficients, though we could not get it to perfectly
satisfy either independence or separation. 
This is to be expected because this regulariser does not encode these criteria
exactly in its formulation. 
This is further demonstrated in its inflection away from satisfying separation
for larger values of $\regf$.
The convex `individual' regulariser also cannot satisfy these criteria for
the same reason, though it seems to be encoding a fairness criterion
quite dissimilar from either separation or independence.
The linear version of LSPC does not perform well on this dataset at all,
even though it is trying to satisfy these criteria.
The reason for this behaviour is a systemic bias in its estimation
of conditional entropy arising from its lower modelling capacity.
In this case it pays off to take the (small) computational cost
of the quadratic basis version of the LSPC approximation for 
and improved conditional entropy approximation.

In the LSAC dataset for the most part we can see similar performance
to the CC dataset. The notable exception to this is now the linear
LSPC regulariser has sufficient modelling capacity to (mostly) satisfy
the separation and independence fairness criteria.

We also ran timing comparisons between the different regularisation methods,
results shown in Figure~\ref{fig:reg_time}. These results have been
generated using increasing subset sizes of the LSAC dataset.
The bars indicate the variation of time with respect to changing $\regf$ ---
typically larger values were used so we emphasis the optimisation time of
``fair'' models.
All other parameters were held constant.
We believe these are good representations of performance since we implemented
all methods, and used exactly the same base linear model code.
These results are in-line with our expectations, as the~\cite{berk_2017}
regularisers scale as \bigo{\N^2}, whereas our LSPC is simply a 
linear classifier, and so scales as \bigo{\N}.

We believe these results support our claims of LSPC being an efficient and
effective means of approximating a conditional entropy based fairness
regulariser. In the next experiment section we will further explore the
efficacy of LSPC in approximating our normalised MI measures, $\nmi$.

\subsection{Quality of Mutual Information Approximations}%
\label{sub:exp_mi}
The fact that the accuracy$/$fairness trade-off curves of the proposed method are comparable to those of a state-of-the-art method suggests that our approximations to mutual information despite fast may be indeed reasonable. In this section we empirically investigate this hypothesis. 

\begin{figure}[t!]
    \centering
    \begin{subfigure}{0.32\textwidth}
        \includegraphics[width=\textwidth]{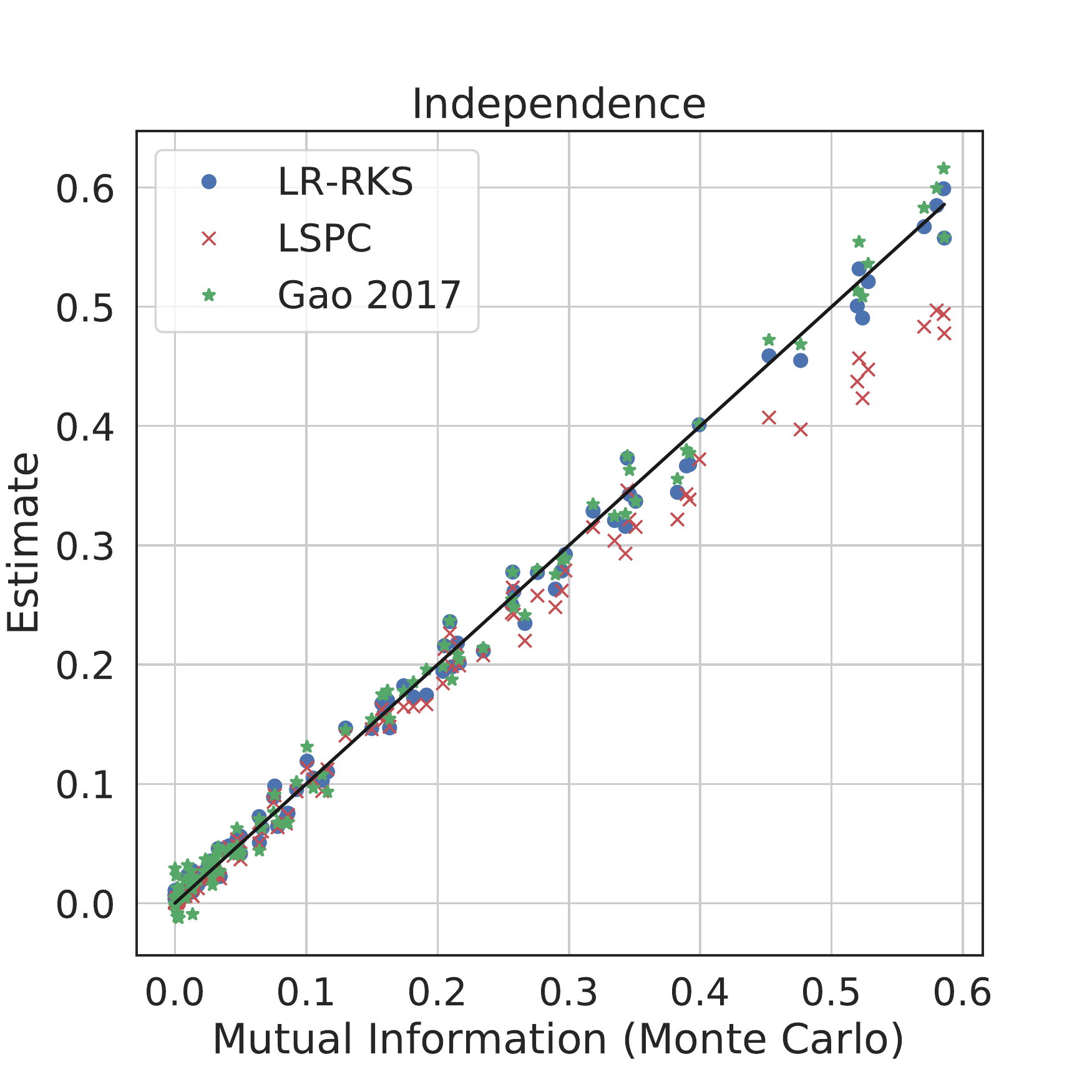}
        \caption{\label{ind_mi_acc}}
    \end{subfigure}\\
    \begin{subfigure}{0.32\textwidth}
        \includegraphics[width=\textwidth]{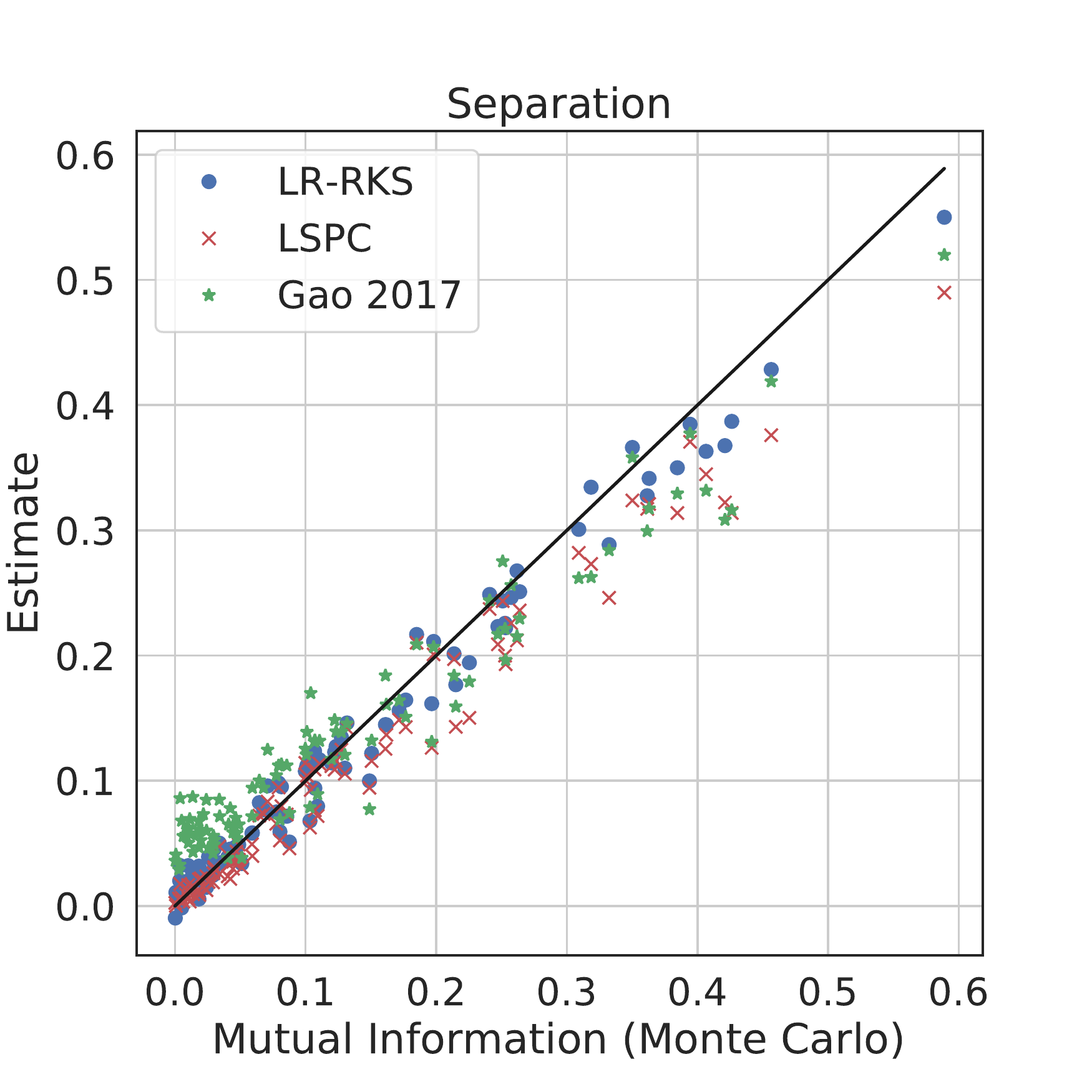}
        \caption{\label{sep_mi_acc}}
    \end{subfigure}\\
    \begin{subfigure}{0.32\textwidth}
        \includegraphics[width=\textwidth]{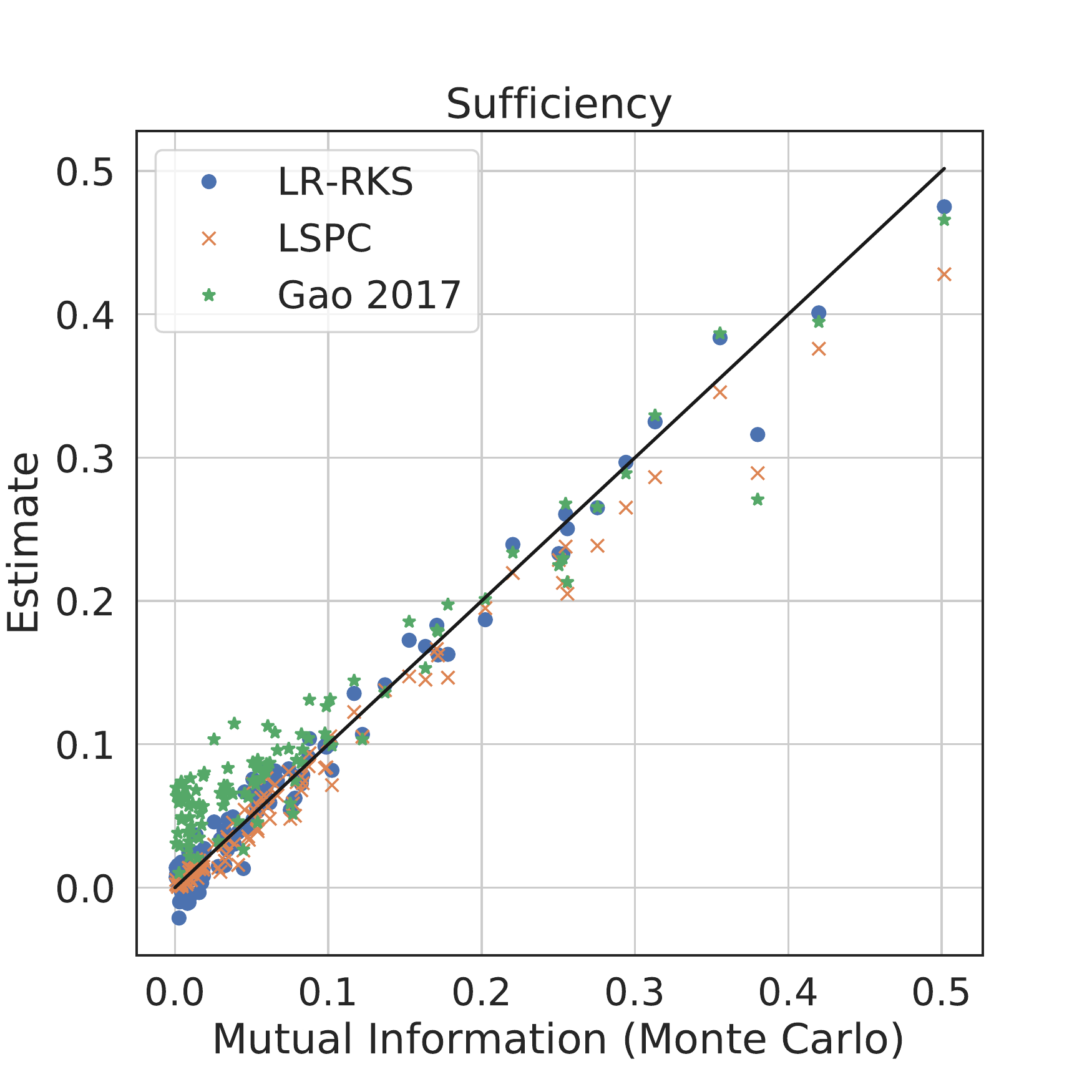}
        \caption{\label{suf_mi_acc}}
    \end{subfigure}
\caption{Comparison of MI estimators against the Monte Carlo approximation for a parameterised joint distribution.}
\label{fig:mc_comp}
\vspace{-5mm}
\end{figure}

\begin{figure}[]%
 \centering
  \includegraphics[width=0.7\linewidth]{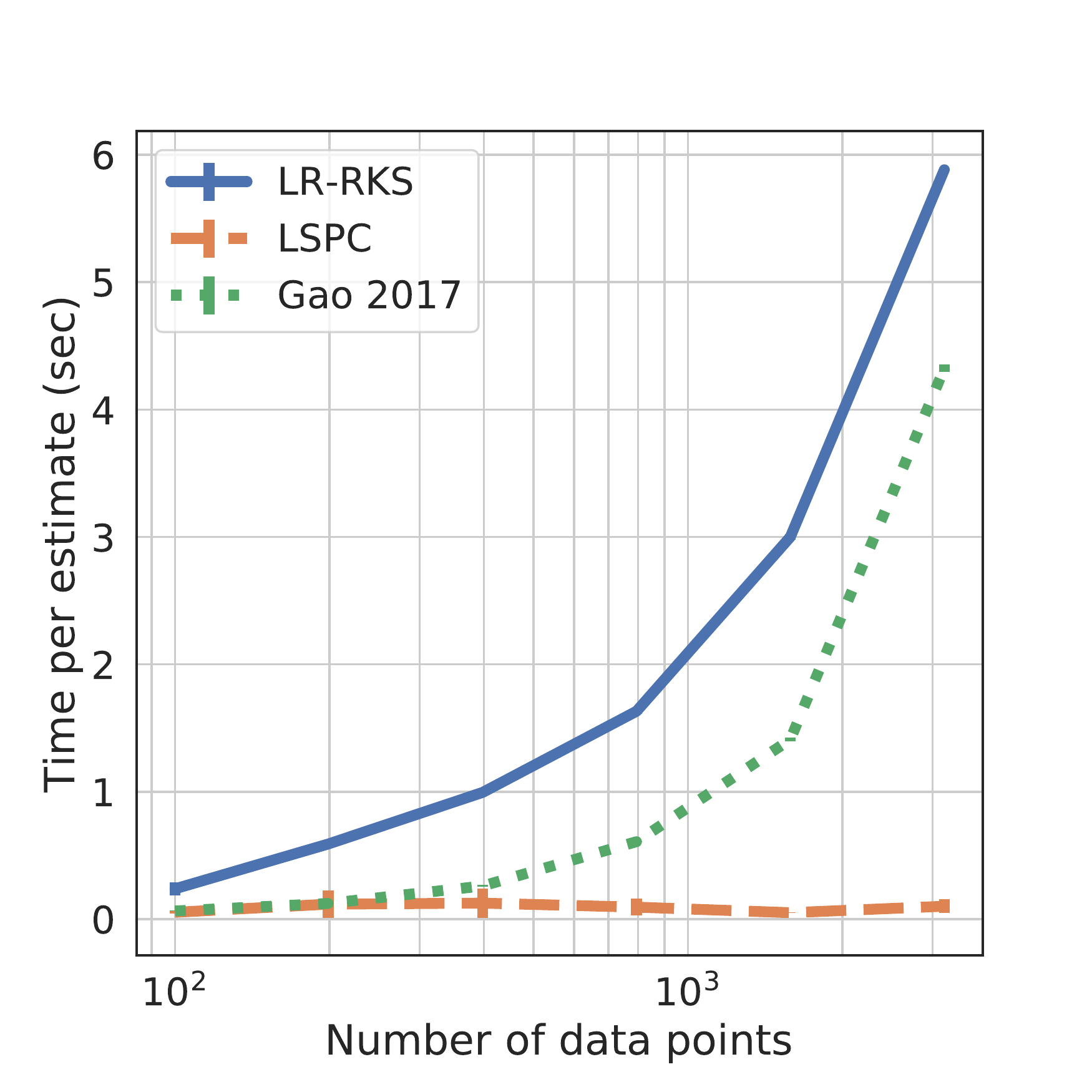}
\caption{Run time comparison of mutual information approximation methods.}
\label{fig:timings_mi}
\end{figure}

To estimate the accuracy of the proposed MI estimators, we created a parameterised joint probability distribution over $\Y$, $\Sc$ and $\A$ in which the true mutual information could be independently quantified to a high precision using Monte Carlo integration.
We used a Bernoulli distribution to model a binary sensitive attribute, $\A$, which then conditioned a multivariate Gaussian distribution that represented the target values and regressor scores $\Y$ and $\Sc$, respectively.

%(Fig.~\ref{fig:synth_data_model}).

%\begin{figure}[h!]
% \label{fig:synth_data_model}
% \centering
% \tikz{
% \node[latent] (A) {$\A$};
% \node[latent,right=of A,xshift=0cm,fill] (ys) {$\Y, \Sc$}; %
% %\draw (0,1.69) circle(.36cm); 
% \edge{A}{ys}
% }
%\end{figure}

Varying the mean and covariance of the conditional Gaussian distribution enabled us to generate joint distributions that exhibited potentially problematic characteristics from a fairness perspective. These include differences in base-rates, regressor accuracy and score across the protected classes.

Figure~\ref{fig:mc_comp} illustrates the performance of each MI estimator when
compared against the Monte Carlo approximation of mutual information over 100
different scenarios (the Monte Carlo approximation is our gold standard in this
evaluation). LR-RKS and Gao 2017 are a logistic regressor with Fourier features
(kitchen sinks) sinks~\cite{rahimi_2008} and the KSG estimator proposed
in~\cite{gao_2017}, respectively. These are used in \S\ref{sub:exp_reg} as
independent estimates of MI because an analytical joint distribution was not
available. LSPC refers to the our fast approximation of MI described in
\S\ref{sub:fairreg}. We can see from the figure that the quality of our LSPC
approximation is indeed reasonable --- although not as good as for the most
expensive methods. Further refining the quality of this approximation without
significantly affecting its efficiency is left as a direction for future work.

Fig.~\ref{fig:timings_mi} shows that despite the fact that the quality of our mutual information approximation is similar to that of the other methods, its computational efficiency is significantly higher. This is the key justification behind the proposed approach.

\section{Conclusion}

In this paper we have studied the problem of learning a regression model under
group fairness constraints as defined by the fairness criteria of independence,
separation and sufficiency. Deviations from perfect alignment with each of
these criteria can be quantified in terms of (conditional) mutual information,
which motivates the use of these quantities as regularisers in order to
trade-off accuracy and fairness in an regularised empirical risk minimisation
framework. General approximations for mutual information can be computationally
expensive, so we devised a fast approximation procedure particularly suited for
the mixed categorical$/$continuous distributions arising due to protected
classes being discrete and predicted outputs continuous. We have shown
experimentally that the proposed approximation to mutual information, although
much faster than alternatives, is not significantly less accurate. As a result,
when we then incorporated these approximations as regularisers in a regression
setting, we obtained comparable accuracy$/$fairness trade-offs to a
state-of-the-art method --- while being significantly more computationally
efficient. 

%\section*{Acknowledgements}
%
%Withheld for review.


\bibliography{drfair,ethicalml_bib}
\bibliographystyle{icml2020}

\appendix
\section{Entropy Regularisers --- Extended Results}

The main paper proposes fairness regularisers promoting statistical separation, independence or sufficiency in a regression setting.
The approach employs probabilistic classification to estimate conditional probabilities of the protected attributes.

This means that a classification model is nested inside the regression model's training objective.
The classifier's \emph{trained} predictive outputs, and their gradients, need to be efficiently computed at each step of the training optimisation for this approach to be computationally tractable.

For this purpose we employed the LSPC classifier, a least-squares approximation of a logistic regressor~\cite{sugiyama_efficient_2010, sugiyama_superfast_2010}. 
In our application, this model closely approximates the performance of the same objective using a standard logistic regressor, while offering a faster closed form expression for the trained classifier's predictive probabilities.
Evidence of this statement was deferred to this supplementary material.

Here we include experimental results obtained using a standard logistic regression classifier inside our fairness regulariser.
The logistic regression model is using the same quadratic feature expansion as our LSPC-quad implementations, and retrained at every evaluation of the regressor loss using nested numerical optimisation.
Relevant regression objective gradients were inferred using knowledge that the classifier's objective gradient was zero.

\begin{figure*}[bth]%
\centering
\begin{subfigure}[t]{\linewidth}
  \centering
  \includegraphics[width=0.9\linewidth]{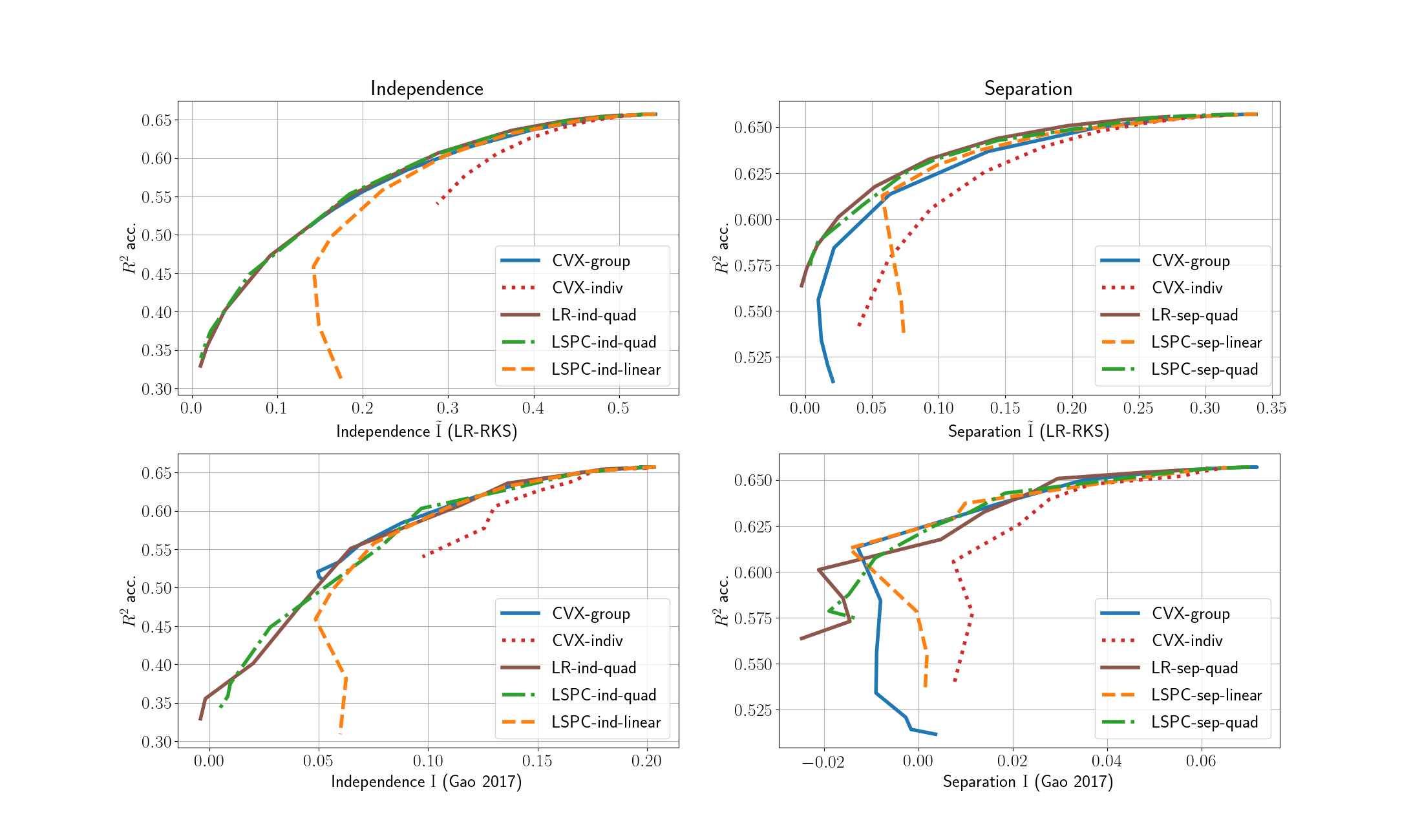}
  \caption{Communities and Crime dataset.\label{fig:reg_comm_app}
  }
\end{subfigure} \\
\begin{subfigure}[t]{\linewidth}
  \centering
  \includegraphics[width=0.9\linewidth]{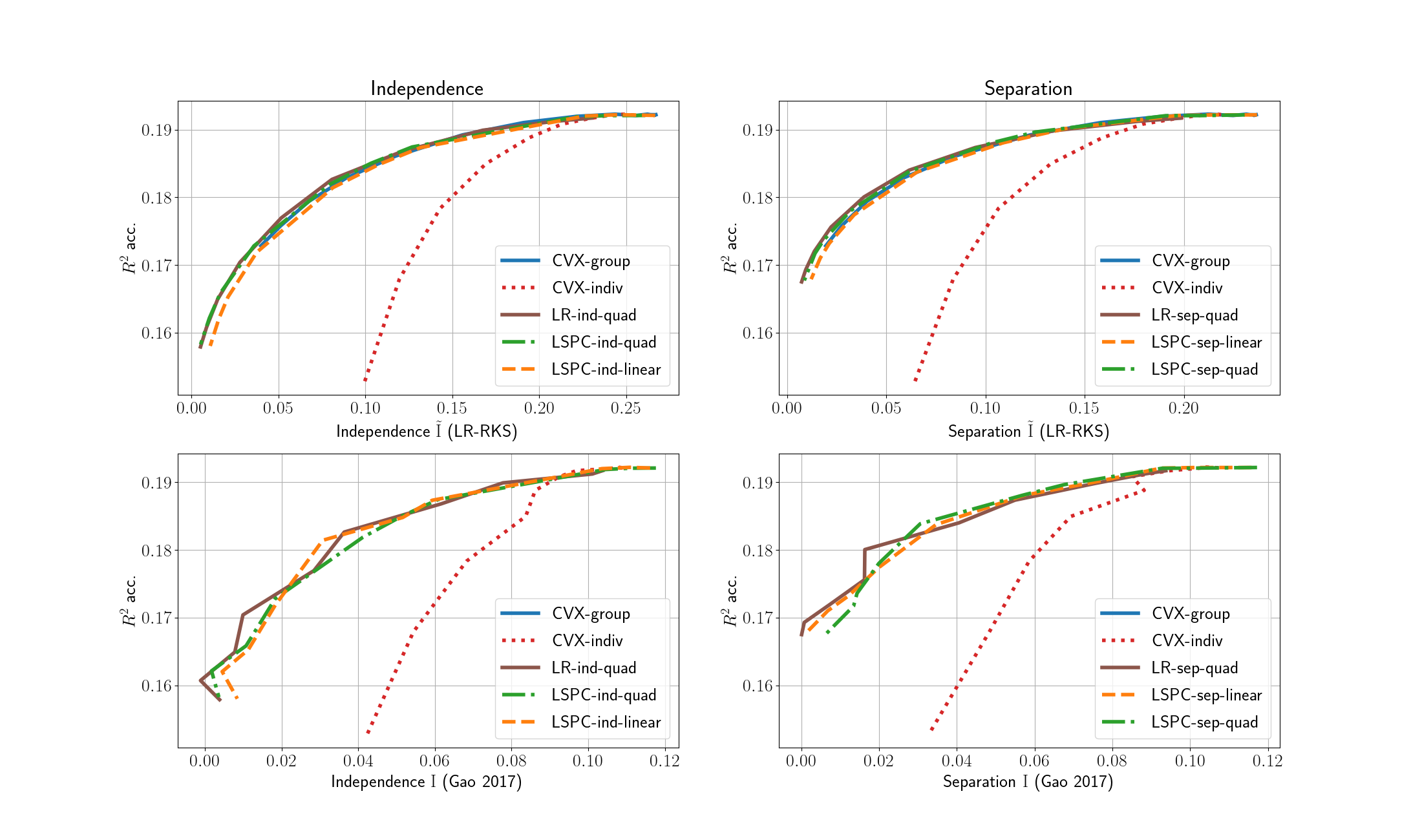}
  \caption{LSAC dataset.\label{fig:reg_lsac_app}}
\end{subfigure}
\caption{
  Efficiency frontier, comparing our regulariser as proposed with a fast approximate classifier (LSPC) to our regulariser using a standard logistic regression model on the same features (LR).
  As detailed in the main paper, the CVX-group and CVX-indiv are the group and individual fairness regularisers proposed in~\citet{berk_2017}, and LSPC are our proposed implementation, with `ind' and `sep' denoting the independence and separation objectives, and `linear' and `quad' denoting a linear or feature-cross basis.
  The frontier is plotted in terms of $R^2$ accuracy, and Independence and Separation using normalised conditional mutual information (using both using probabilistic classification (LR-RKS), and mixed discrete-continuous MI estimation procedure of~\citet{gao_2017} as detailed in the main paper).
}
\end{figure*}

\begin{figure}[ht]%
  \includegraphics[width=1.05\linewidth]{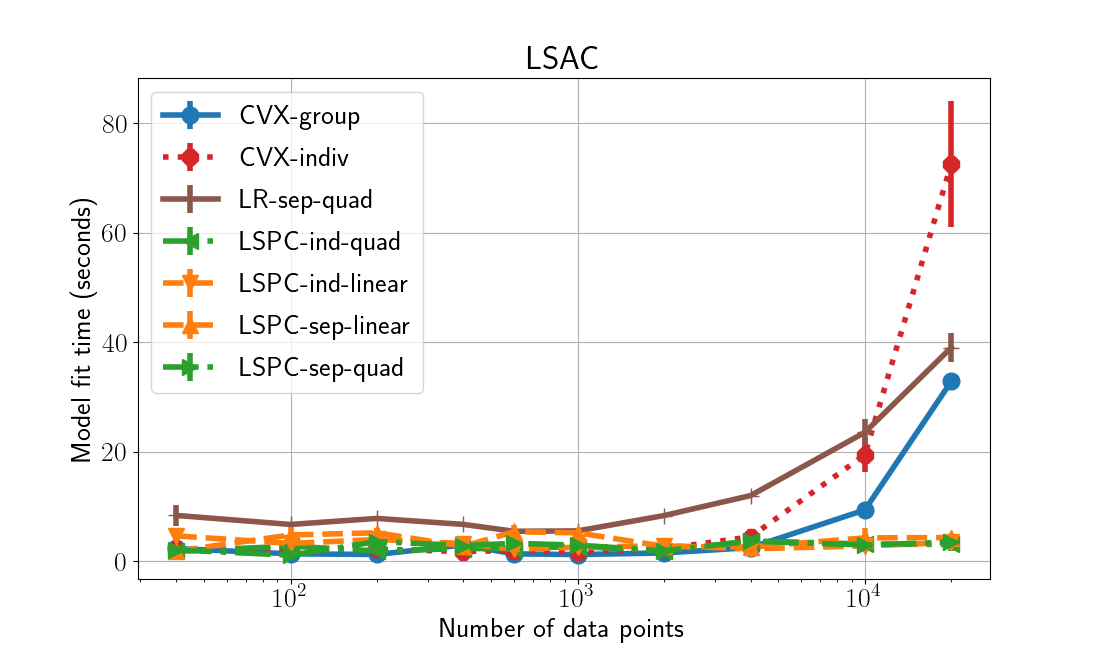}
  \hspace{-.05\linewidth}
  \caption{
    \label{fig:timing_app}
    Regressor training times as a function of the size of the training dataset using each fairness regressor.
    % The LR-sep-quad method for separation) is compared to the methods from 
    % the main paper (including our and the regularisers proposed in~\citet{berk_2017}).
    The proposed LSPC-sep-quad (employing a least squares probabilistic classifier) is a fast approximation of LR-sep-quad (employing a logistic regression classifier), and scales better with dataset size.
    The x-axis denotes the subset size of the LSAC dataset used to time the model fitting, and the bars depict the range of times for a sweep of regulariser strengths,~$\regf$.
    % We note that LSPC-sep-quad is our fast (and equivalently performant) approximation of LR-sep-quad.
  }%
  \label{fig:reg_time_app}
\end{figure}

As shown in Figures~\ref{fig:reg_comm_app} and~\ref{fig:reg_lsac_app}, over our experimental datasets the LSPC-quad indeed offers equivalent predictive performance to the full logistic regressor that it approximates. 
On the other hand, the timing plots in Figure~\ref{fig:timing_app} show that the least squares approximation is significantly faster and scales better with respect to dataset size.

\end{document}